\pdfoutput=1

\documentclass[11pt]{article}

\usepackage{acl}

\usepackage{times}
\usepackage{latexsym}

\usepackage[T1]{fontenc}

\usepackage[utf8]{inputenc}

\usepackage{microtype}
\usepackage{times}
\usepackage{url}
\usepackage{latexsym}
\usepackage{amssymb}

\usepackage{titlesec}
\usepackage{multirow}
\usepackage{subfig}
\usepackage{mathtools}
\usepackage{colortbl}
\usepackage{float}
\usepackage{xcolor}
\usepackage{enumitem}
\usepackage[capitalize]{cleveref}

\usepackage{booktabs}
\usepackage[font=small,labelfont=bf,tableposition=top]{caption}
\usepackage{adjustbox}
\DeclareCaptionLabelFormat{andtable}{#1~#2  \&  \tablename~\thetable}
\usepackage[noend]{algpseudocode}
\usepackage{algorithm}
\usepackage{algorithmicx}
\usepackage{arydshln}

\usepackage{pifont}
\newcommand{\cmark}{\ding{51}}%
\newcommand{\xmark}{\ding{55}}%

\usepackage{xspace}

\newcommand{\methodname}{ICXML\xspace}

\title{Generate and Rerank: An In-context Learning Framework for Zero-shot Extreme Classification}
\title{\methodname: An In-Context Learning Framework for\\Zero-Shot Extreme Multi-Label Classification}

\author{Yaxin Zhu \and Hamed Zamani\\
  Center for Intelligent Information Retrieval \\
  University of Massachusetts Amherst \\
  \texttt{\{yaxinzhu, zamani\}@cs.umass.edu} 
  }

\begin{document}
\maketitle
\begin{abstract}

This paper focuses on the task of Extreme Multi-Label Classification (XMC) whose goal is to predict multiple labels for each instance from an extremely large label space. While existing research has primarily focused on fully supervised XMC, real-world scenarios often lack supervision signals, highlighting the importance of zero-shot settings. Given the large label space, utilizing in-context learning approaches is not trivial. We address this issue by introducing In-Context Extreme Multi-label Learning (\methodname), a two-stage framework that cuts down the search space by generating a set of candidate labels through in-context learning and then reranks them. Extensive experiments suggest that \methodname advances the state of the art on two diverse public benchmarks. 
\end{abstract}

\section{Introduction}
Extreme Multi-Label Classification (XMC) deals with the classification of instances into a set of relevant labels from a large label set \cite{bhatia2015sparse,mittal2021eclare,dahiya2023ngame}. It finds applications in various domains, including text categorization \cite{chalkidis2019large}, recommendation systems \cite{agrawal2013multi}, image tagging \cite{Mittal22}, and so on. Unlike conventional multi-label classification, where the number of labels is relatively small, XMC involves an exponentially larger label space, e.g., in the $10^6$ magnitude. This poses significant computational and modeling challenges.

While existing research has primarily focused on supervised XMC, real-world applications often encounter challenges in obtaining complete supervision signals. Scenarios arise during test sessions when new labels emerge without any assigned input instances \cite{gupta2021generalized}, or when both instances and labels are available, but the corresponding relations between them are unknown \cite{xiong2021extreme}. This task, which is the focus of this paper, is called zero-shot XMC.

Zero-shot XMC can be seen as a \emph{retrieval} problem, where the test instance is considered as the query and candidate labels are retrieved in response to the given input. Methods based on lexical matching, such as TF-IDF \cite{Salton1988TFIDF} and BM25 \cite{Robertson1995OkapiBM25}, and semantic matching, such as dense retrieval models \cite{hofstatter2021efficiently,karpukhin-etal-2020-dense}, can be adopted for this task. State-of-the-art approaches for zero-shot XMC, such as RTS \cite{zhang2022structural} and MACLR \cite{xiong2021extreme}, also belong to this category. A major shortcoming with these approaches is that there is little lexical or semantic overlap between the test instance (i.e., queries) and the label space (i.e., documents). One may argue that large language models (LLMs) can be used to generate labels for each test input. However, the labels that LLMs generate may not be in the acceptable label set and unlike conventional classification tasks, the label set is too large to be given to the LLMs in their prompt. Therefore, using LLMs for this task is either impractical or extremely expensive.

In this work, we put together the benefits of both retrieval- and generation-based approaches by introducing \methodname -- a two-stage framework designed for zero-shot XMC. In the first stage - generate, \methodname enriches the intrinsic capabilities of large language models by generating and/or retrieving demonstrations in a zero-shot manner. The obtained outputs are generated by prompting the model with the help of a support set of generated demonstrations. Subsequently, the generated outputs are adjusted to align with the label space, resulting in a condensed shortlist of candidate labels. 
In the second stage - rerank, we leverage the capabilities of LLMs to perform multi-label classification by reintroducing the refined candidate label shortlist along with the test instance as input. This approach capitalizes on the language model's inherent potential to handle multiple labels concurrently, thereby augmenting its performance in the context of extreme multi-label classification tasks.

In summary, our main contributions are three fold:
\begin{enumerate}[leftmargin=*,itemsep=0mm]
    \item Introducing a two-stage framework for zero-shot XMC, involving generation-based label shortlisting and label reranking.
    \item Advocating for a generation-based approach to yield high-quality input-label pairs instead of retrieval-based. This method also addresses the challenges posed by the absence of specific input scenarios, ensuring robustness across diverse contexts. 
    \item Advancing state of the art in zero-shot XMC on two public benchmarks, i.e., LF-Amazon-131K and LF-WikiSeeAlso-320K,  and providing detailed analysis for a deeper understanding of model performance. We show that \methodname performs effectively even without reliance on an input corpus -- a collection of input candidates that is used by state-of-the-art baselines \cite{xiong2021extreme,zhang2022structural}. 
\end{enumerate}

Our implementation scripts and codes are publicly available for research purposes at \url{https://github.com/yaxinzhuars/icxml}. 

\section{Related Work}
\subsection{Extreme Multi-label Classification}
Extreme classification refers to the task of making predictions over vast label spaces, typically comprising thousands to millions of classes, with multiple correct classes assigned to each instance \cite{agrawal2013multi,bhatia2015sparse,liu2017deep,jiang2021lightxml,dahiya2021deepxml,mittal2021eclare,dahiya2023ngame}. In this context, \cite{gupta2021generalized} framework focuses on predicting unseen labels, while \citet{zhang2022metadata} handles instances where no labels are observed. Prior work by \citet{simig2022open} explores the generation of labels usingLLMs for this specific task. Additionally, \citet{xiong2021extreme} investigated a generalized zero-shot setting where no annotations are available. These research endeavors contribute to the advancement and understanding of extreme classification, addressing challenges related to unseen labels, missing label information, and generalized zero-shot scenarios \cite{zhang2022structural,aggarwal2023semsup}. The zero-shot setting has found applicability in various real-world scenarios including cold start recommendation tasks, and is mainly solved with dependency on large-scale training by creating pseudo annotations. In our work, we propose a fully zero-shot setting and aim to tackle it through the utilization of in-context learning.

\subsection{In-Context Learning}
The scaling of model size and corpus size has led to notable advancements in LLMs \cite{brown2020language,chowdhery2022palm}, enabling them to demonstrate remarkable ICL capabilities \cite{wei2022emergent}. These models have showcased their ability to effectively learn from a limited number of examples provided within the context. The research community has witnessed the emergence of numerous studies focusing on the analysis and enhancement of demonstrations in ICL \cite{wei2022chain, fu2022complexity}. \cite{liu2021makes,rubin2021learning,luo2023dr,chen2022decoupling,ram2023context,cheng2023uprise} explored the retrieval of influential demonstrations from a training corpus to provide effective guidance. Additionally, \cite{lyu2022z, chen2023self} proposed the generation of pseudo demonstrations to tackle zero-shot scenarios. 
In addition, a plenty of studies have focused on investigating the ranking ability of models in various tasks characterized by a vast search space. These tasks encompass areas such as information retrieval \cite{shen2023large, gao2022precise}, reranking \cite{sun2023chatgpt,ma2023zero}, and recommendation systems\cite{hou2023large}.
Motivated by these advancements, our objective is to identify an optimal ICL method suitable for extreme multi-label classification, considering the specific requirements and challenges of this task. Through our research, we aim to contribute to the development of effective and efficient ICL techniques that can address the complexities of extreme multi-label classification.

\begin{figure*}
    \centering
     \includegraphics[width=\linewidth]{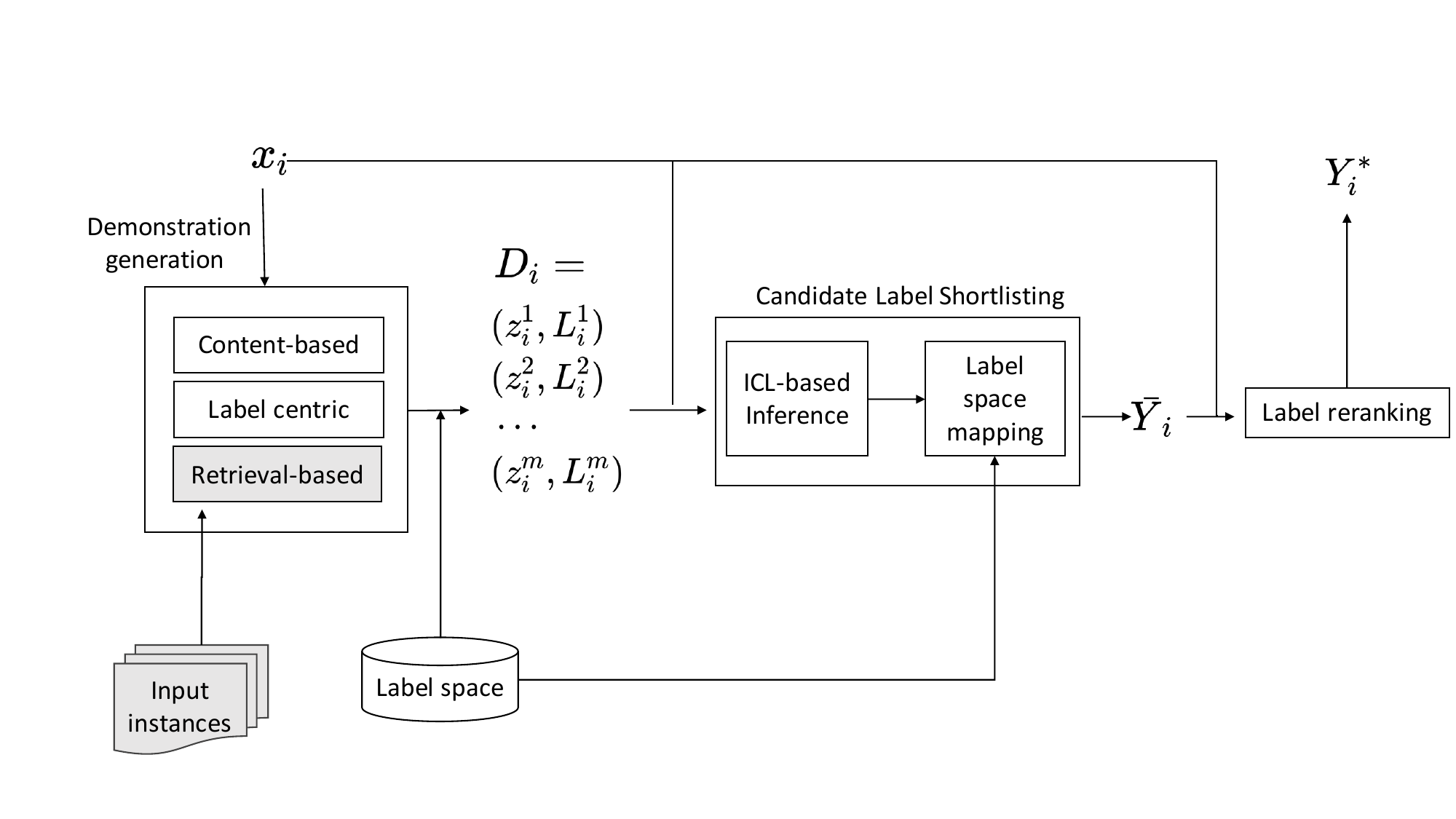}
    \caption{An illustration of the proposed generate-rerank framework. For a given test input \( x_i \), we generate demonstrations \( D_i \) to facilitate ICL-based shortlisting. Subsequently, this shortlisted set \( \bar{Y}_i \) is provided to LLM for listwise re-ranking, culminating in the final results \( Y^*_i \).}
    \label{fig:model}
\end{figure*}

\section{Problem Formulation}
Let $\mathcal{X}$ and $\mathcal{Y}$ respectively denote the input and output spaces. In this work, we focus on text data, thus each $x \in \mathcal{X}$ is an unstructured text and each $y \in \mathcal{Y}$ is represented by a short text description. Given the focus on XMC, the output space is extremely large, e.g., $|\mathcal{Y}| \sim 10^6$. The goal is to map each input $x$ to a small subset of labels $Y \subset \mathcal{Y}$.

Following \citet{xiong2021extreme, zhang2022structural}, we also consider a scenario where some input instances, called the input corpus, are available, however the mapping $\{(x_i, Y_i)| x_i \in \mathcal{X}_{\text{train}}, Y_i \subseteq \mathcal{Y}\}$ is not available for training.

\section{The \methodname Framework}

Benefiting from the high performance and rich knowledge encoded within LLM parameters, we can easily generate a set of labels by describing classification tasks and inputs using prompts. However, for a classification task with a predetermined label set, this approach results in uncertain outcomes in the absence of guidance from few-shot example pairs, referred to as demonstrations \cite{reynolds2021prompt, razeghi2022impact}. While common approaches involve incorporating the label candidate list into the prompt, this approach becomes impractical or extremely expensive when faced with an extreme label space, e.g., a label space of $10^6$ magnitude.


To address this issue, we propose a two-stage framework illustrated by \ref{fig:model}. The first stage is generate, including demonstration generation (Section \ref{sec:demogen}) and candidate shortlisting (Section \ref{sec:shortlisting}) in the figure. We perform in-context learning using LLM $\phi$ to generate labels, and the pseudo demonstrations for this stage is generated by $\phi$ using a prompt-guided approach. The second stage is rerank, where we utilize another prompt-guided method and $\phi$ for selecting top labels from candidate labels as decribed in Section \ref{sec:rerank}.

\begin{figure*}
    \centering
     \includegraphics[width=\linewidth]{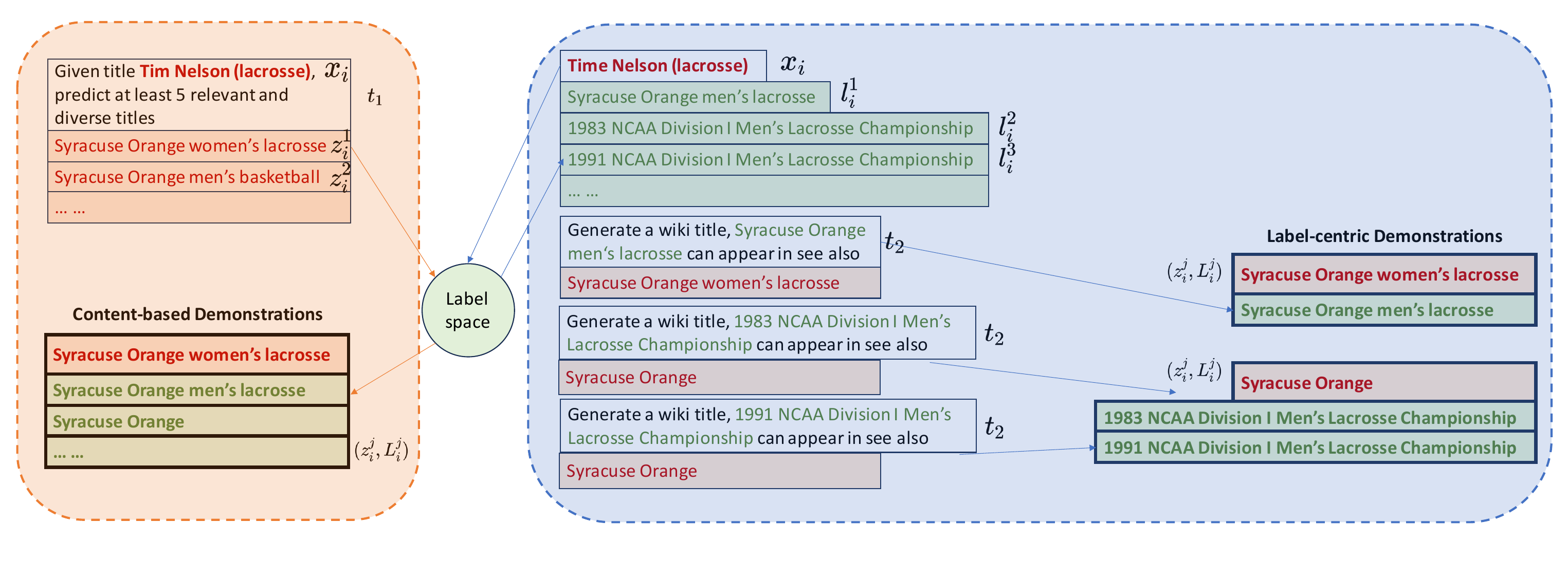}
    \caption{Pipeline of content-based and label-centric approaches for demonstration generation stage. In the content-based paradigm, demonstrations are generated through LLM, producing inputs denoted as $z_i^j$, followed by the selection of corresponding outputs $L_i^j$ from the label space. Conversely, in the label-centric paradigm, demonstration outputs $l_i^j$ are initially retrieved from the label space, subsequently leading to the determination of corresponding inputs $z_i^j$.}
    \label{fig:4.1}
\end{figure*}


\subsection{Demonstration Generation} \label{sec:demogen}
For an effective in-context learning performance, it is essential for demonstrations to encompass both the inherent correlation between the input text and the task label, as well as external knowledge that facilitates the model's learning process in relation to the input text. We propose two different strategies, illustrated in Figure \ref{fig:4.1} to achieve this goal.  

\medskip

\noindent\textbf{Content-based Demonstration Generation: }
To embody a blend of external knowledge and inherent correlation, the content-based approach first generates relevant and diverse demonstration inputs based on the \emph{test input content}. Each input is then linked to a label that exhibits the inherent correlation.

For each test input $x_i \in \mathcal{X}$, the LLM $\Phi$ is employed to generate a set of $m$ demonstration inputs, denoted as ${Z}_i = \{z_i^1, z_i^2, \cdots, z_i^m\}$, where $z_i^j \sim \Phi(\textsc{Prompt}(x_i, t_1))$, $t_1$ is the corresponding task description.\footnote{See Appendix \ref{instructions} for detailed task descriptions for all 4 different tasks} For each $z_i^j \in {Z}_i$ and each label $l \in \mathcal{Y}$, the zero-shot retriever $\theta$ computes a score function as follows: $\text{score}(l) = \theta (l| z_i^j)$. The top $n$ labels, denoted by ${L}_i^j = \arg \mathrm{top} \mathit{n} \{\text{score}(l)|l \in \mathcal{Y}\}$ are selected based on their scores to form the demonstration set: 
\begin{equation}
{D}_i = \{(z_i^1, {L}_i^1), (z_i^2, {L}_i^2), \cdots, (z_i^m, {L}_i^m)\}.
\end{equation}

\noindent\textbf{Label-centric Demonstration Generation:}
The label-centric approach pursues an inherent-external trajectory by first mapping the test input to the label space, capturing labels with high correlation. It then generates demonstration inputs based on these labels.

For each test instance $x_i \in \mathcal{X}$, the zero-shot retriever $\theta$ identifies the top $n$ labels, denoted by ${L}_i = \{l_i^1, l_i^2, \ldots, l_i^n\}$, where $L_i = \arg \mathrm{top} \mathit{n} \{\theta(l|x_i)|l \in \mathcal{Y}\} $. For each label $l_i^j \in {L}_i$, a pseudo input text is generated: $z_i^j \sim \Phi(\textsc{Prompt}(l_i^j, t_2))$. When duplicate input texts arise, the labels are merged into a label list, denoted by $\mathbf{l}_i^j$. To be consistent with content-based method, let $L_i^j = \{l_i^j\}$, the demonstration set is then constructed as follows:
\begin{equation}
{D}_i = \{ (z_i^1, {L}_i^1), (z_i^2, {L}_i^2), \ldots, (z_i^m, {L}_i^m)\},\end{equation}
where $m$ is the final size of the grouped label list, and $l_i^j$ corresponds to the duplicate input $z_i^j$.


\subsection{Candidate Label Shortlisting}
\label{sec:shortlisting}
\noindent\textbf{ICL-based Inference}
After the pseudo demonstration sets ${D}_i$s are constructed, we integrate them with each test input $x_i$ in the prompt, guiding the few-shot learning process of the language model $\Phi$. Consequently, $\Phi$ generates a $k$-sized set of labels
$\hat{Y}_i = \{y_{i,1}, y_{i,2}, \cdots, y_{i,k}\}$ , for each $y_{i,j} \in \hat{Y}_i$,
\begin{equation} y_{i,j} \sim \Phi(\textsc{Prompt}(x_i, {D}_i, t_3))\end{equation}
where $y_{i,j}$s are the labels produced by the language model, and $t_3$ denotes the task description. The generation of these labels encapsulates the model's prediction based on both the original input and the pseudo demonstration set.

\noindent\textbf{Label Space Mapping.}\label{sec:label_space_adaption}
After inference, we leverage textual semantic matching techniques to establish connections between the generated text and the corresponding labels in the label space. This step is critical for transforming the raw output from the language model into structured labels.

For each generated label $y_{i,j} \in \hat{Y}_i$, we use the zero-shot retriever $\theta$ to fetch the top $s$ labels from the label set $\mathcal{Y}$ that possess the highest semantic similarity with $y_{i,j}$. This set, denoted by $\bar{Y}_{i,j}$, is defined as:
\begin{equation}
\bar{Y}_{i,j} = \{\bar{y} = \arg\mathrm{top}\mathit{s} \theta (y| y_{i,j}), y \in \mathcal{Y}\}\end{equation} 

Finally, we obtain a shortlist
\begin{equation} \bar{Y}_i = \bigcup_{j} \bar{Y}_{i,j} \end{equation}
for each test instance $x_i$. Through this process, we map the generated labels to the label space while simultaneously expanding them to a desirable size for in-context learning-based multi-label classification. 

\subsection{Label Reranking}
\label{sec:rerank}

With the obtained shortlist, our approach effectively contracts the search space for labels, recasting the problem into a standard multi-label classification task. To benefit from this formulation, we feed the whole shortlist into $\Phi$:
\begin{equation}
\begin{aligned}
{Y}_i^*  \sim\Phi(\textsc{Prompt}(x_i, \bar{Y}_i, t_4)), 
& {Y}_i^* \subseteq \bar{Y}_i
\end{aligned}
\end{equation}

Here, a prompt $\textsc{Prompt}(x_i, \bar{Y}_i, t_4)$, steers the LLM to select the most suitable set of labels. The set of labels chosen, denoted by ${Y}_i^*$, serves as the final prediction in our approach.


\section{Expanding \methodname by Utilizing an Input Corpus}
Our methodology adopts a novel approach that avoids the use of training samples, setting it apart from conventional models. This design choice aligns with zero-shot learning paradigms where, although a corpus of training instances $\mathcal{X}_{\text{train}}$ and their corresponding labels may be available, they are not utilized in a paired manner. 
our framework maintains a degree of adaptability and can be readily extended to a more flexible setting by substituting demonstration generation with demonstration retrieval, when $\mathcal{X}_{\text{train}}$ is available. 

For each test input $x_i \in \mathcal{X}$, we select the top $m$ neighbor instances from $\mathcal{X}_{\text{train}}$ as follows:

\begin{equation}
\begin{aligned}
{Z}_i = \{z | z = \arg\mathrm{top}\mathit{m} \theta (z| x_{i}),  \\| {Z}_{i} | \leq m, z\in \mathcal{X}_{\text{train}}\}
\end{aligned}
\end{equation}

Subsequently, for each $z_i^j \in {Z}_i$, we proceed with the methodology delineated in Section \ref{sec:demogen}, titled ``Content-based demonstration generation'', to construct ${D}_i$. 

\begin{table}[t]
\centering
\resizebox{1\linewidth}{!}{
\begin{tabular}{lccc}
\hline
\textbf{Dataset} & \textbf{$|X_{train}|$} & \textbf{$|X_{test}|$} & \textbf{$|Y|$} \\
\hline
LF-Amazon-131K & 294,805 & 134,835 & 131,073\\ 
LF-WikiSeeAlso-320K & 693,082 & 177,515 & 312,330 \\\hline
\end{tabular}
}
\caption{Data statistics. The size of training instances, test instances and label space are presented.}
\label{tab:data}
\end{table}

\begin{table*}[t]
\centering
\resizebox{\textwidth}{!}{
\begin{tabular}{@{}lcccclcccclcccclcccc@{}}
\toprule
 & & \multicolumn{8}{c}{\textbf{LF-Amazon-131K}}&\multicolumn{8}{c}{\textbf{LF-WikiSeeAlso-320K}}   \\
\cmidrule{3-5} \cmidrule{7-10}\cmidrule{12-14}\cmidrule{16-19}
 & $\mathcal{X}_{\text{train}}$ & P@1 & P@3 & P@5 && R@1 & R@3 & R@5 & R@10 && P@1 & P@3 & P@5 && R@1 & R@3 & R@5 & R@10  \\
\midrule
TF-IDF & \xmark & 0.124 & 0.115 & 0.091 && 0.069 & 0.181 & 0.231 & 0.293 && 0.107 & 0.089 & 0.071 && 0.059 & 0.130 & 0.165 & 0.216 \\
 BM25 & \xmark & 0.174 & 0.118 & 0.088 && 0.100 & 0.185 & 0.224 & 0.268  && 0.185 & 0.120 & 0.090 && 0.101 & 0.175 & 0.210 & 0.254 \\
 TAS-B & \xmark & 0.135 & 0.123 & 0.096 && 0.081 & 0.203 & 0.255 & 0.313 && 0.237 & 0.161 & 0.125 && 0.131 & 0.238 & {0.292} & {0.365} \\
 MACLR & \cmark & 0.181 & 0.154 & 0.119 && 0.104 & 0.244 & 0.304 & 0.373 && 0.163 & 0.135 & 0.108 && 0.097 & 0.204 & 0.254 & 0.321 \\
 RTS & \cmark & 0.187 & 0.153 &  {0.120} && 0.106 & 0.242 & 0.304 & 0.382 && 0.186 &  {0.151} &  {0.121} && 0.108 &  {0.227} &  {0.283} & 0.354\\
 free generation & \xmark &  0.171 & 0.110 & 0.084 && 0.097 & 0.177 & 0.219 & 0.274 && 0.246 & 0.156 & 0.116 && 0.133 & 0.228 & 0.271 & 0.327\\\hdashline
 \methodname{} - content & \xmark & 0.225 & 0.148 & 0.109 &&  \textbf{0.141} &  {0.266} &  {0.320} & 0.349 && 0.241 & 0.150 & 0.109 && 0.128 & 0.201 & 0.242 & 0.301 \\ 
 \methodname{} - label & \xmark &  \textbf{0.234} &  \textbf{0.160} &  \textbf{0.119} && 0.134 & 0.250 & 0.300 &  {0.357}  &&  \textbf{0.278} &  \textbf{0.169} &  \textbf{0.125} &&  {0.149} &  \textbf{0.246} & \textbf{0.290} & 0.356\\
  \methodname{} - retrieval & \cmark &  {0.220} &  {0.155} & 0.115 &&  {0.135} &  \textbf{0.279} & \textbf{0.342} &  \textbf{0.404} & &  {0.252} & 0.141 & 0.105 &&  \textbf{0.152} & 0.225 & 0.256 &  \textbf{0.361}  \\
\bottomrule
\end{tabular}
}
\caption{Experimental results obtained by the proposed approach and the baselines. The highest number in each column is bold-faced. For proposed approach, we are using GPT-3.5 for in-context learning.}
\label{tab:result}
\end{table*}

\section{Experiments}

\subsection{Data}
We evaluate the effectiveness of our approach on the following large-scale datasets: LF-Amazon-131K in item recommendation domain and LF-WikiSeeAlso in Wikipedia articles title tagging domain, where 131K and 320K denote the size of label space \cite{Bhatia16}. The dataset statistics are presented in Table \ref{tab:data}, showcasing the characteristics of each dataset. These benchmark datasets are widely used in evaluation of zero-shot extreme classification settings.

\subsection{Baselines}
We compare our approach with lexical matching based methods, soft semantic matching based methods, pseudo pretraining based methods and naive zero-shot in-context learning without demonstration augmentation:

\noindent\textbf{Lexical Matching}: TF-IDF is a powerful sparse lexical matching technique that matches input tokens to the nearest labels based on similarity in terms of bag-of-words representation \cite{Salton1988TFIDF}. BM25 is a term-based ranking model that scores documents based on their term frequencies and document lengths. \cite{Robertson1995OkapiBM25}.

\noindent\textbf{Soft Matching}: The recent development of language models and pre-training + fine-tuning paradigms has paved the way for zero-shot learning using soft matching techniques. Among these models, TAS-B has emerged as an effective and lightweight pre-trained bi-encoder, demonstrating strong generalization capabilities \cite{hofstatter2021efficiently}. TAS-B is trained using dual supervision from a cross-encoder model and ColBERT on the MS MARCO dataset \cite{nguyen2016ms}.

\noindent\textbf{MACLR \cite{xiong2021extreme}} was trained with pseudo positive pairs constructed from TF-IDF.

\noindent\textbf{RTS \cite{zhang2022structural}} proposed a self-supervised auxiliary task for contrastive representation learning that enables end-to-end training. 

\noindent\textbf{Free Generation of LLM}: The LLM is provided solely with the test input and the task objective, which encompasses elements such as task description, output constraints, among others, to generate a prediction. The prediction is derived by adapting labels from the actual label space (refer to Section \ref{sec:label_space_adaption}), guided by a heuristic methodology which selects the nearest adaptation in the order of generated raw labels (refer to Section \ref{sec:ab}) 

\subsection{Experimental Setup}
For the LLM $\Phi$, we use OpenAI's GPT-3.5 API in main experiments, while GPT-4 is included in ablation study on a small subset. The LLM was called with a temperature hyperparameter set to 0.0. Instructions are provided in appendix.
For semantic matching $\theta$, we use TAS-B model as our semantic macher, so that we can benefit from its knowledge acquired by the pretraining on MS MARCO.
For content-based generation, we set $m=5,n=5$.
For label-centric generation, we set $n=30$.
For inference, we set $k=10,s=10$.
Following the setup of \cite{xiong2021extreme}, we use precision and recall as evaluation metrics.

\begin{figure*}[h]
\begin{center}$
\begin{array}{cc}
\includegraphics[width=0.45\textwidth]{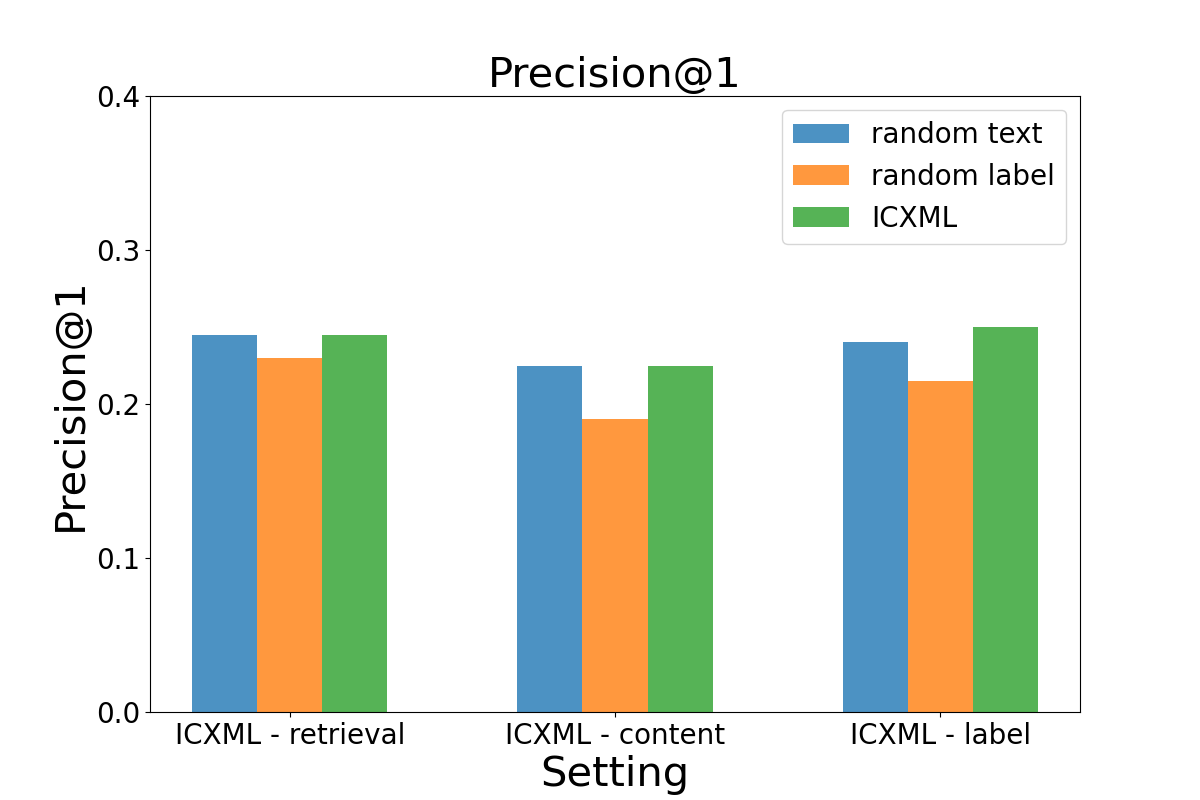}&
\includegraphics[width=0.45\textwidth]{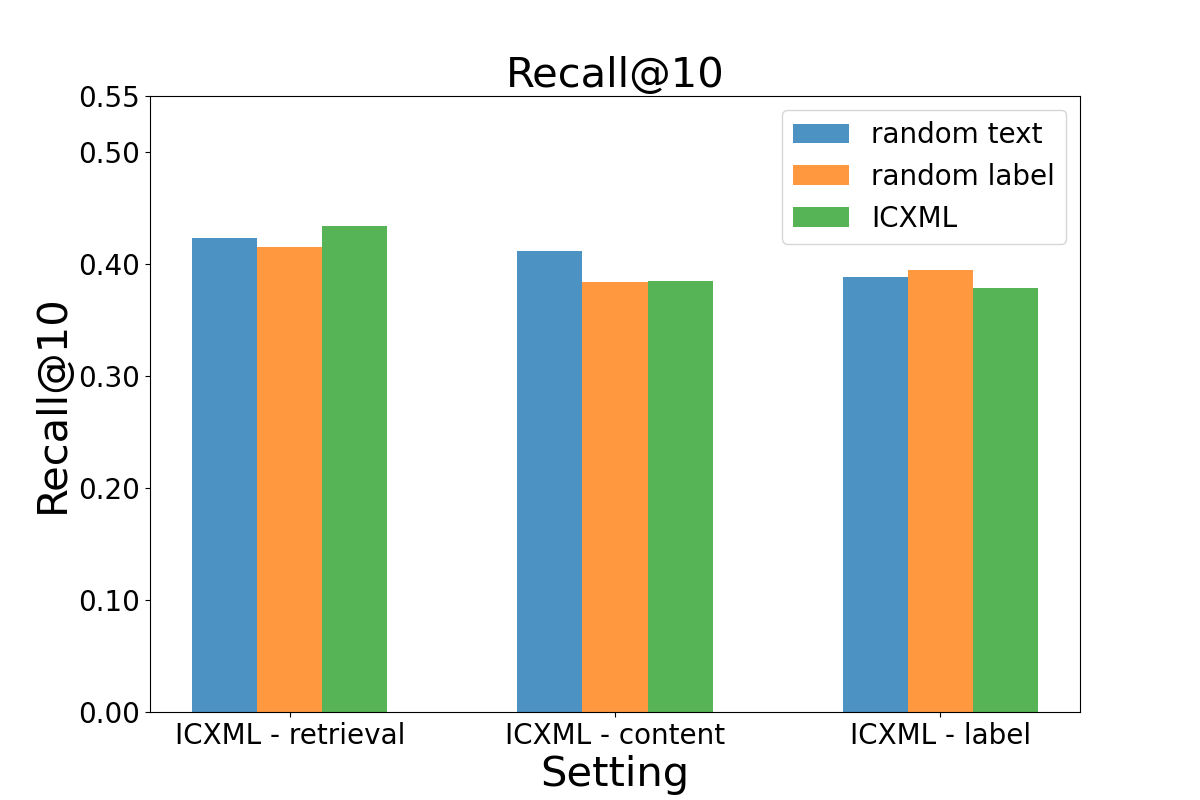}
\end{array}$
\end{center}
\caption{Results of different demonstration construction strategies on 200 samples from LF-Amazon-131K. }
\label{fig:demo_ret}
\end{figure*}

\subsection{Main Results}
The main experimental results of extreme classification are presented in Table \ref{tab:result}. In this table, the various \methodname{} suffixes denote distinct methods used for constructing demonstrations, and others are the baselines. Here, content-based and label-centric (denoted as ``content'' and ``label'') are two different strategies of our method, while demonstration retrieval results (denoted as "retrieval") are recorded to investigate the differential impacts of demonstration retrieval and generation on the in-context learning framework. In general, our method substantially outperforms all baselines when applied to the LF-Amazon-131K dataset, demonstrating a significant enhancement in performance. For LF-WikiSeeAlso-320K dataset, it surpasses the highest performing baseline, soft matching (tas-b), by 3.9\% and 1.8\% in terms of P@1 and R@1 respectively. However, when evaluating a longer result list, it does not measure up to the performance levels achieved by soft matching. The underperformance might be due to the large label space of 320K, which is significantly larger than 131K. This imposes challenges for the generation-based label space adaptation in terms of fully capturing the label distribution compared to retrieval-based methods. Despite this, we executed experiments on a small subset of the test set using GPT-4 in section \ref{sec:ab}. It was found that this underperformance could be completely mitigated by employing GPT-4 instead of GPT-3.5 for the reranking component of our approach.

It is also observed that for LF-Amazon-131K dataset, the performance of the label-centric approach and the content-based approach comparably equivalent, whereas for LF-WikiSeeAlso-320K, the label-centric approach demonstrates a clear superiority. This discrepancy can be ascribed to the differing degrees of correlation between labels and primary identifiers (product names for LF-Amazon-131K and wiki page titles for LF-WikiSeeAlso-320K). The Amazon dataset demonstrates a stronger association between labels and product names, while in the case of the WikiSeeAlso, the labels appear to have a more substantial dependence on the actual content. Under our experimental framework where only product names or wiki page titles are generated, the relevance of each title-adapted pseudo label is comparatively low. Interestingly, even when provided with access to the input corpus and the ability to construct demonstrations through the retrieval of existing input sources, the performance of demonstration retrieval does not surpass that of demonstration generation. It exhibits better performance in terms of recall@10, but precision@1, intriguingly, is even superior in the case of generation. The reason could be diversity discrepancy between generated and retrieved inputs and noise in pairing fixed inputs with fixed label space. To gain deeper insights into these observed performance variations, we conducted an extensive analysis in Section \ref{sec:ab}.

\subsection{Ablation Study} \label{sec:ab}
In this section, we study the top 1 and top 10 performance differences. Also, 
we conduct comprehensive ablation analyses to discern the contribution of each component in \methodname{}. These evaluations are performed on a sample comprising 200 instances from both test dataset. We use label-centroid generation.
In our ablation study, we answer the following empirical research questions:

\begin{table*}[t]
\resizebox{1.0\textwidth}{!}{
\begin{tabular}{@{}llccclcccclccclcccc@{}}
\toprule
 && \multicolumn{8}{c}{\textbf{LF-Amazon-131K}} &  \multicolumn{8}{c}{\textbf{LF-WikiSeeAlso-320K}}\\
\cmidrule{3-5} \cmidrule{7-10} \cmidrule{12-14} \cmidrule{16-19} 
 generation and shortlisting && P@1 & P@3 & P@5 && R@1 & R@3 & R@5 & R@10 && P@1 & P@3 & P@5 && R@1 & R@3 & R@5 & R@10   \\
\midrule
 free generation && 0.160 & 0.128 & 0.087 && 0.099 & 0.235 & 0.257 & 0.324 && 0.250 & \textbf{0.174} & \textbf{0.136} && 0.124 & 0.244 & 0.304 & 0.346 \\\hdashline
 TAS-B  && 0.190 & 0.131 & 0.094 && 0.115 & 0.236 & 0.290 & 0.354 && 0.270 & 0.172 & \textbf{0.136} && 0.144 & \textbf{0.262} & \textbf{0.314} & \textbf{0.381}\\
 free + TAS-B results as hint && 0.175 & 0.117 & 0.084 && 0.105 & 0.200 & 0.234 & 0.293 && 0.225 & 0.145 & 0.106 && 0.118 & 0.210 & 0.233 & 0.322 \\\hdashline
 ICL &&  \textbf{0.220} & \textbf{0.140} &  \textbf{0.107} &&  \textbf{0.135} &  \textbf{0.252} & \textbf{0.312} & \textbf{0.361} &&  \textbf{0.310} & 0.173 & 0.124 && \textbf{0.164} & 0.250 & 0.276 & 0.353  \\ 
\bottomrule 
\end{tabular}
}
\caption{Ablation study of generate stage on sampled LF-Amazon-131K and LF-WikiSeeAlso-320K datasets of size 200. ``free generation'' is the original in-context learning configuration with a reranking stage, ``TAS-B'' is applying reranking stage on top 100 results of TAS-B, ``free + TAS-B results as hint'' is using these top 100 labels as hint in free generation prompt. ``ICL" is our proposed method. Reranking techniques are all based on LLM (GPT-3.5)}
\label{tab:ab-short}
\end{table*}

\begin{table*}[t]
\resizebox{1.0\textwidth}{!}{
\begin{tabular}{@{}llccclcccclccclcccc@{}}
\toprule
 && \multicolumn{8}{c}{\textbf{LF-Amazon-131K}} &\multicolumn{8}{c}{\textbf{LF-WikiSeeAlso-320K}}  \\
\cmidrule{3-5} \cmidrule{7-10} \cmidrule{12-14} \cmidrule{16-19}
 reranking && P@1 & P@3 & P@5 && R@1 & R@3 & R@5 & R@10 && P@1 & P@3 & P@5 && R@1 & R@3 & R@5 & R@10  \\
\midrule
  heuristic && 0.160 & 0.128 & 0.087 && 0.099 & 0.236 & 0.257 & 0.324 && 0.285 & 0.168 & \textbf{0.127} && 0.140 & 0.239 & \textbf{0.283} & 0.328\\
  monoT5 && 0.180 & 0.117 & 0.089 && 0.119 & 0.215 & 0.272 & 0.345 && 0.220 & 0.137 & 0.105 && 0.111 & 0.190 & 0.229 & 0.281\\
  LLM &&  \textbf{0.220} & \textbf{0.140} &  \textbf{0.107} &&  \textbf{0.135} &  \textbf{0.252} & \textbf{0.312} & \textbf{0.361} &&  \textbf{0.310} & \textbf{0.173} & 0.124 && \textbf{0.164} & \textbf{0.250} & 0.276 & \textbf{0.353}  \\ 
\bottomrule 

\end{tabular}
}
\caption{Ablation study of rerank stage on sampled LF-Amazon-131K and LF-WikiSeeAlso-320K datasets of size 200. ``heuristic'' can be regarded as natural results without reranking. ``monoT5'' is a pretrained ranking model. ``LLM'' is listwise reranking based on LLM.}
\label{tab:ab-reranking}
\end{table*}

\begin{table*}[t]
\resizebox{1.0\textwidth}{!}{
\begin{tabular}{@{}lllccclcccclccclcccc@{}}
\toprule
 &&& \multicolumn{8}{c}{\textbf{LF-Amazon-131K}} &\multicolumn{8}{c}{\textbf{LF-WikiSeeAlso-320K}}  \\
\cmidrule{4-6} \cmidrule{8-11} \cmidrule{13-15} \cmidrule{17-20} generate &
 reranking && P@1 & P@3 & P@5 && R@1 & R@3 & R@5 & R@10 && P@1 & P@3 & P@5 && R@1 & R@3 & R@5 & R@10  \\
\midrule
  GPT-3.5 & GPT-3.5 &&  {0.220} & 0.140 &  {0.107} &&  {0.135} &  {0.252} & {0.312} & 0.361 &&  \textbf{0.310} & 0.173 & 0.124 && \textbf{0.164} & 0.250 & 0.276 & 0.353  \\ 
  GPT-3.5 & GPT-4 && {0.220} & \textbf{0.142} & 0.105 && {0.135} & 0.243 & 0.310 & \textbf{0.397} && 0.295 & \textbf{0.202} &\textbf{0.148} && 0.148 & \textbf{0.282} & \textbf{0.332} & \textbf{0.427} \\
  Llama 2 & RankVicuna && \textbf{0.225} & 0.141 & \textbf{0.108} && \textbf{0.140} & \textbf{0.256} & \textbf{0.316} & {0.365} && 0.265 & 0.151 & 0.111 && 0.130 & 0.206 & 0.238& 0.317\\
\bottomrule 

\end{tabular}
}
\caption{Ablation study of different language models on sampled LF-Amazon-131K and LF-WikiSeeAlso-320K datasets of size 200. RankVicuna is fine-tuned on Llama 2 for zeroshot listwise reranking.}
\label{tab:ab-model}
\end{table*}

\begin{table*}[t]
\centering
\resizebox{\textwidth}{!}{
\begin{tabular}{lccc}
\hline
\textbf{Model}  & TAS-B & free generation & ICXML\\\hline

\textbf{Candidate 1} & Bethany & Norway Lutheran Church & \underline{\textbf{National Register of Historic Places listings in Latah County, Idaho}} \\
\textbf{Candidate 2} & Norway Lutheran Church &  Church of Norway & Carpenter Gothic\\
\textbf{Candidate 3} & Church of Norway & Churches in Norway & National Register of Historic Places\\
\textbf{Candidate 4} & Churches in Norway & Norwegian Church, Cardiff & List of Lutheran churches\\
\textbf{Candidate 5} & Norwegian Church, Cardiff & The Norwegian Lutheran Church in the United States & List of Idaho counties\\
\bottomrule
\end{tabular}
}
\caption{Top 5 candidates for query "Bethany Memorial Chapel" in WikiSeeAlso task across three settings, with "National Register of Historic Places listings in Latah County, Idaho" as the only gold label.}
\label{tab:example}
\end{table*}

\noindent \textbf{RQ1}: \textit{How is the effect of different demonstration construction strategies?}

We evaluate three distinct demonstration construction strategies on a subset of 200 instances extracted from LF-Amazon-131K test dataset. The strategies employed were as follows: retaining the original demonstrations, replacing the input text with random words, and replacing the paired labels with random labels. As presented in Table \ref{fig:demo_ret}, it becomes evident that the most substantial decrease in Precision@1 performance occurs when the paired labels are replaced. This observation underscores the critical role played by label space coverage in the preparation of demonstrations. The results further imply the flexibility and efficacy of \methodname{} in handling XMC challenge with or without an input corpus.

\noindent \textbf{RQ2}: \textit{How is the effect of generate and shortlisting?}

To discuss the effect of generate and shortlisting component, we simply modify the free generation setup, transitioning from an approximation of generated raw labels to our unique generate-rerank paradigm, which allows for an expanded scope of label mappings. Under this scheme, we regard the enlarged set via label space mapping of raw labels as a condensed candidate list, from which the top 10 are selected via listwise reranking based on GPT-3.5.  In Table \ref{tab:ab-short}, ``free generation'' and ``TAS-B'' denotes further reranking from the top 100 results derived from the no-demonstration prompting configuration or TAS-B respectively.
In an effort to rigorously evaluate the quality of the generated demonstration, we conducted an additional experimentation aiming at understanding the performance influence of input text that is exclusively generated by the LLM and moves beyond the boundaries of the accessible corpus. For this test, we used a prompt that was filled with pseudo labels serving as hints but did not include any pseudo pairs. Results are denoted as ``free + TAS-B results as hint''. 

Comparing the results of all experiments, it is evident that our approach exhibits superior performance, affirming the effectiveness of demonstration generation. By feeding all soft matching results to LLM as hint to generate label candidate shortlist, ``free + TAS-B results as hint'' underperforms compared to the direct utilization of these results as candidates, highlighting the key role of external knowledge and inherent correlation between input and label conveyed by generated demonstrations.

\noindent \textbf{RQ3}: \textit{How is the effect of label reranking?}

For this research question, we keep generated candidates frozen, but apply different strategies to produce the final top 10 answers. The strategies include:
\noindent\textbf{Heuristic}: This strategy is incorporated within free generation configuration to opt for results from the generated label list that are most proximate to the mapped label space. For the $i$th instance, represent mapped labels based on generated labels $Y_i$ as $\{y_{i,1}^1, \cdots, y_{i, 1}^{10}, \cdots y_{i, k}^{10}\}$, where $\{y_{i,1}^1, \cdots, y_{i, 1}^{10}\}$ are top 10 neighbor labels identified by an zero-shot retriever. Simply rerank $Y_i$ with a heuristic rule: Specifically, the ranking is carried out in the order of the generated labels, where the nearest neighbor within the label candidate set is chosen. If this nearest neighbor has already been arranged, defer to the second nearest one. This procedure continues in a similar manner for subsequent generated labels. The final result should be $\{y_{i,1}^1, y_{i,2}^1, \cdots, y_{i, 10}^{10}, \cdots y_{i, k}^{10}\}$.
\noindent\textbf{MonoT5}: a ranking model which fundamentally leverages T5 architecture to calculate
    the relevance score, and is fine-tuned on MS MARCO passage dataset. Within the context of this comparison, we employ the monoT5-3B model variant for our analyses and evaluations.
\noindent\textbf{LLM}: Our strategy introduced in section \ref{sec:rerank}.

Experimental results of the three strategies are presented in Table \ref{tab:ab-reranking}. The observed enhancement when transitioning from heuristic to other strategies implies that increasing the number of neighbors incorporated within the label space mapping step can result in more correct labels being added to the shortlist. This suggests that a robust reranking approach has potential utility and validates the effectiveness of our generate-rerank framework. By broadening the candidate set, our framework provides an enriched space for accurate label selection, demonstrating its value in complex label space adaptations. Further improvement from monoT5 to LLM indicate the LLM's strong ability of multi-choice selection and reranking. 

\noindent \textbf{RQ4}: \textit{To what extent does the performance of our method generalize to different models?}

GPT-4 is confirmed to be significantly superior in terms of ranking performance \cite{sun2023chatgpt}. 
Due to the extensive size of the test sets, we confined our LLM-based listwise reranking using GPT-4 to a small subset of the dataset. The findings demonstrate that GPT-4 excels in reranking tasks, particularly with the WikiSeeAlso dataset. Here, our method, \methodname{}, consistently surpasses the baselines. These results underscore the potential capability of LLMs to adapt to extensive label spaces, thereby illustrating their utility in extreme classification scenarios.

Furthermore, to address the concerns raised about the reproducibility and robustness of our framework, particularly regarding the use of large black box models such as GPT 3.5/4.0, we have expanded our research to include experiments with the more recent and open-sourced large language model, Llama 2 \cite{touvron2023llama}. In the generate and shortlisting stage, we used vanilla Llama 2 to construct demonstrations and generate candidate shortlist. In the reranking stage, we used RankVicuna \cite{pradeep2023rankvicuna}, a model that has undergone instruction tuning and further distillation of knowledge derived from GPT-4's listwise reranking outcomes based on Llama 2. This was implemented to replicate our initial zeroshot listwise setup. The results, as shown in the Table \ref{tab:ab-model}, offer insights into the model-agnostic nature of our method and its efficacy across different large language model platforms. 

\subsection{Case study}
Table \ref{tab:example} presents the outcomes for a specific query in the WikiSeeAlso task. Notably, despite being the gold label and frequently encountered in related contexts, "National Register of Historic Places listings in Latah County, Idaho" failed to appear in the candidate lists generated by TAS-B or free generation methods. This observation underscores the challenges inherent in accurately capturing relevant content solely through these approaches.

When employing ICXML, a notable improvement is observed. Interestingly, the absence of the gold label from the initial candidate lists is compensated by the generated demonstration input: "National Register of Historic Places: Bethany Memorial Chapel." This demonstrates the potential of in-context learning to enhance the relevance and inclusivity of candidate recommendations by leveraging contextual information, thereby enriching the overall user experience and retrieval accuracy in the WikiSeeAlso task.

\section{Conclusions and Future Work}
In conclusion, this paper addressed the challenges of Extreme Multi-Label Classification (XMC) in real-world scenarios with limited supervision signals. We proposed the \methodname{} framework to handle this setting without reliance on input text and pretraining. Experimental results demonstrated the effectiveness of our approach in improving the performance of XMC and its various zero-shot settings. Our research contributes to the advancement of XMC by offering new insights and methodologies for addressing real-world challenges.

For future work, an interesting direction would be to evaluate the adaptability of \methodname{} across diverse domains and multi-modal data. Understanding how the model behaves with different domain-specific terminologies and when combined with visual or auditory data will be crucial. Also, Combining \methodname{} with other state-of-the-art XMC techniques might offer synergistic benefits. Exploring hybrid models can potentially unlock new efficiencies and improved performance.

\section*{Risks and Limitations}
Like other LLM based works, one of the risks of this work is ethical and bias considerations: Any biases present in the training data of ChatGPT will influence the generated labels. Without appropriate checks, these biases might amplify or result in misleading labels, especially in sensitive areas. 

Furthermore, the adaptability of \methodname{} to data from diverse domains and in multi-modal formats is an area yet to be explored thoroughly. The behavior of the language model may vary based on the distinctiveness and intricacies of specific domain terminology or when integrating visual cues. Addressing these variations will be crucial for \methodname{} to be universally applicable.


\section*{Acknowledgments}
This work was supported in part by the Center for Intelligent Information Retrieval and in part by NSF grant \#2106282. Any opinions, findings and conclusions or recommendations expressed in this material are those of the authors and do not necessarily reflect those of the sponsor.

\bibliography{anthology,custom}
\bibliographystyle{acl_natbib}

\appendix

\section{Appendix: Instructions}\label{instructions}
See instructions on the next page.

\begin{table*}
    \begin{tabular}{lll}
        $t_1$: generate demonstration input based on test input &&   \\
    \end{tabular}
\end{table*}

\begin{table*}
    \begin{tabular}{lll}
        **Product title**: & \texttt{test input title}  \\
        **Task**: & Please predict at least 5 relevant and diverse Amazon products titles. \\
        **Format**: & ["title1", "title2", "title3", "title4", "title5"], do not say any word or explain. \\
        **Product Description**: & \texttt{test input description}\\
    \end{tabular}
\end{table*}

\begin{table*}
    \begin{tabular}{lll}
        **Wiki title**:  & \texttt{test input title} \\
        **Task**: & Please generate at least 5 relevant and diverse Wikipedia page titles.  \\
        **Format**: & ["title1", "title2", "title3", "title4", "title5"], do not say any word or explain. \\
        **Wiki content**: & \texttt{test input title} \\
    \end{tabular}
\end{table*}

\begin{table*}
    \begin{tabular}{lll}
        $t_2$: generate demonstration input based on label input &&   \\
    \end{tabular}
\end{table*}

\begin{table*}
    \begin{tabular}{ll}
        \multicolumn{2}{l}{For an Amazon product recommendation task, } \\
        **Product title**: & \texttt{test input title}  \\
        **Candidate labels**: & \texttt{retrieved labels} \\
        **Task**: & For each label, guess an input title. \\
        **Format**: & ["title1", "title2", "title3", "title4", "title5"], each title is a guess based on a \\
        & candidate label, title1 is a guess for first label, and so on. Only output one list \\
        & and the list should be of size 30. do not explain or say anthing. \\
    \end{tabular}
\end{table*}

\begin{table*}
    \begin{tabular}{ll}
        As 'See Also' pages of & \texttt{test input title} \\
        There's a list of Wikipedia page titles: & \texttt{retrieved labels} \\
        **Task**: & For each page, generate a "See also" page title.  \\
        **Format**: & ["title1", "title2", "title3", "title4", "title5"], each title is \\
        & a guess based on a candidate label, title1 is a guess for \\
        & first label, and so on. Only output one list and the list \\
        & should be of size 30. do not explain or say anthing. \\
    \end{tabular}
\end{table*}

\begin{table*}
    \begin{tabular}{lll}
        $t_3$: Inference &&   \\
    \end{tabular}
\end{table*}

\begin{table*}
    \begin{tabular}{lll}
        **Product title**: & \texttt{demonstration input title}  \\
        **Relevant product**: & \texttt{corresponding labels}  \\
        \multicolumn{2}{l}{... ...} & \\
        **Task**: & Please predict at least 10 relevant products for a new Amazon product title: \\
        &\texttt{test input title} \\
        **Product Description**: & \texttt{test input description}\\
        **Format**: & Only output titles with line break, do not include anything else. 
    \end{tabular}
\end{table*}

\begin{table*}
    \begin{tabular}{lll}
        **Wiki title**: & \texttt{demonstration input title}  \\
        **'See Also' pages**: & \texttt{corresponding labels}  \\
        \multicolumn{2}{l}{... ...} & \\
        **Title**: & \texttt{test input title} \\
        **Content**: & \texttt{test input description}\\
        **Task**: & Generate 'See also' suggestions related to the Wikipedia title \texttt{test input title} \\
        **Format**: & Only output titles with line break, do not include anything else. \\
    \end{tabular}
\end{table*}

\begin{table*}
    \begin{tabular}{lll}
        $t_4$: Reranking &&   \\
    \end{tabular}
\end{table*}

\begin{table*}
    \begin{tabular}{lll}
        **Task**: & Given a query product, select the top 10 most relevant products \\&from a list of candidates. \\
        **Query product title**: & \texttt{test input title}  \\
        **Format**: & A list of integers representing the indices of the top 10 most possible titles. \\&Example: [1, 2, 3, 4, 5, 6, 7, 8, 9, 10] \\
        **Candidates**: &\texttt{label shortlist} \\
        **Product Description**: & \texttt{test input description}\\
    \end{tabular}
\end{table*}

\begin{table*}
    \begin{tabular}{lll}
        **Task**: & From the following candidate list of Wikipedia pages, select top 10 that would \\&be most relevant for the 'See also' section of the given page:
 \\
        **Wiki title**: & \texttt{test input title}  \\
        **Format**: & A list of integers representing the indices of the top 10 most possible titles. \\&Example: [1, 2, 3, 4, 5, 6, 7, 8, 9, 10] \\
        **Candidates**: &\texttt{label shortlist} \\
        **Wiki Content**: & \texttt{test input description}\\
    \end{tabular}
\end{table*}

\begin{table}[htbp]
\centering
\begin{tabular}{|m{2cm}|m{3cm}|m{4cm}|}
\end{tabular}
\label{tab:your-table-label}
\end{table}

\end{document}